\documentclass[letterpaper]{article} 
\usepackage[preprint]{aaai2027}  
\usepackage[hyphens]{url}  
\usepackage{graphicx} 
\usepackage{natbib}  
\usepackage{caption} 
\usepackage{algorithm}
\usepackage{algorithmic}

\usepackage{amsthm}
\usepackage{amsmath}
\usepackage{amssymb}

\usepackage{newfloat}
\usepackage{listings}
\DeclareCaptionStyle{ruled}{labelfont=normalfont,labelsep=colon,strut=off} 
\floatstyle{ruled}
\newfloat{listing}{tb}{lst}{}
\floatname{listing}{Listing}

\usepackage{booktabs}

\title{One Frame, Full Heartbeat: ECG-Free Cardiac Cine MRI Synthesis via Phase-Conditioned Flow Matching}
\author{
    Shiyi Wang\textsuperscript{\rm 1,\rm 2},
    Ruochen Sun\textsuperscript{\rm 2},
    Xiang Li\textsuperscript{\rm 3},
    Peirong Liu\textsuperscript{\rm 1}\corresponding,
    Fangxu Xing\textsuperscript{\rm 2}\corresponding
}
\affiliations{
    \textsuperscript{\rm 1}Department of Electrical and Computer Engineering, Johns Hopkins University, Baltimore, MD, USA\\
    \textsuperscript{\rm 2}Gordon Center for Medical Imaging, Department of Radiology, Massachusetts General Hospital and Harvard Medical School, Boston, MA, USA\\
    \textsuperscript{\rm 3}Center for Advanced Medical Computing and Analysis, Department of Radiology, Massachusetts General Hospital and Harvard Medical School, Boston, MA, USA
}

\begin{document}

\maketitle

\begin{abstract}
Cine cardiovascular magnetic resonance (CMR) analysis  relies on multi-frame sequences capturing the full cardiac cycle. However, standard multi-frame acquisition depends heavily on electrocardiogram (ECG) gating and repeated breath-holds, posing challenges in uncooperative populations, resource-limited settings, and temporally corrupted datasets. Existing methods that synthesize full cardiac sequences
either rely on explicit ECG signals to parameterize myocardium function, or employ
deformable registration without physiological constraints, failing to faithfully
reproduce clinically relevant dynamic metrics such as ejection fraction (EF) and ventricular contraction magnitude.
We present \textbf{PhaseFlow}, a unified generative framework that overcomes both limitations. PhaseFlow estimates a non-linear cardiac phase signal directly from the input sequence via a segmentation-derived left-ventricular (LV) area curve, capturing the asymmetric dynamics of systole and diastole without any ECG dependency. At inference, this phase signal is provided by a pathology-specific template, informing phase-specific frame generation. A rectified flow model conditioned on the phase and slice position synthesizes the full cardiac motion trajectory in the latent space, decoded into a diffeomorphic displacement field that warps end-diastole pixel intensities directly, eliminating the reconstruction blur often accompanying the variational autoencoder. On the ACDC benchmark, PhaseFlow achieves superior physiological fidelity and image realism, with best LV
volume curve $R^2$, structural similarity (SSIM) and generative quality (FID) among all baselines. Ablation studies confirm that each proposed component contributes measurably to the overall performance.
\end{abstract}


\section{Introduction}

Cine cardiovascular magnetic resonance (CMR)  imaging is the clinical gold standard
for quantifying ventricular anatomy and function, enabling precise
physiological measurements throughout the complete cardiac
cycle such as ejection fraction (EF), end-diastolic (ED) and end-systolic (ES)
volumes, myocardial strains, etc.~\cite{bernard2018acdc}.
In standard clinical practice, a full CMR acquisition often requires electrocardiogram (ECG) gating and multi-breath-hold pulse sequences, a
protocol unavailable in retrospective cohort studies, prohibitively costly
in resource-limited settings, challenging for uncooperative population, and inapplicable when only isolated frames
survive archiving.
Therefore, for any specific population with consistent cardiac functional patterns, the problem of synthesizing a \emph{complete, physiologically coherent} CMR
sequence from any specific frame (e.g., the ED frame where the myocardium expands the most) is  both clinically
compelling and technically formidable: it demands recovering the full
spatiotemporal trajectory of the myocardium from purely spatial information,
without any explicit temporal signal.

Prior works have attempted this challenge from three directions, each with
distinct limitations.
\emph{Registration-based} methods~\cite{dragnet, krebs2021} warp a reference
frame via a predicted displacement field, but treat the frame index as a
linear phase proxy, erasing the asymmetric velocity profiles of systole
and diastole~\cite{gptrack} and generalizing poorly under domain
shift~\cite{regmagic}.
\emph{ECG-conditioned} methods~\cite{echopulse, ecgflowcmr} restore temporal
coherence but hard-depend on concurrent ECG acquisition, precisely the signal
prone to absence in retrospective or resource-limited settings.
\emph{Generative} models, namely generative adversarial networks
(GANs)~\cite{goodfellow2014gan, gancmri} and diffusion
models~\cite{ho2020ddpm, song2021score, heartbeat, echodiff_icassp, biomedvid},
optimize pixel-level realism without physiological supervision and thus consistently fail
to reproduce clinically relevant quantities such as EF and left-ventricular (LV)
volumes~\cite{echopulse}; conditional image leakage~\cite{imgleak} further drives
them toward near-static predictions at inference.

We observe that all three paradigms overlook a key signal that is inherently latent in the
CMR sequence itself: the nonlinear cardiac phase, recoverable from the
per-frame LV area curve without any ECG dependency.
Cardiac dynamics are intrinsically asymmetric: systole and diastole span different
fractions of the cycle and proceed at markedly different velocities~\cite{fukuta2008},
and the area swept by the LV myocardium across frames encodes the true temporal
dynamics, capturing the velocity asymmetry that a uniform frame-index assumption fails to.
Beyond this intra-subject asymmetry, different pathological conditions also alter the
characteristics of the phase trajectory in systematic ways~\cite{zheng2019explainable,atehortua2022}:
the relative timing and velocity balance of systole and diastole shift consistently
across disease phenotypes, making the phase curve itself a pathology-level signature.
At inference, when only one single frame is available and no sequence-derived phase
can be computed, a pathology-specific template can replace this estimate, preserving
ECG independence while encoding disease-level motion priors.

We present \textbf{PhaseFlow}, a generative framework that synthesizes full
cardiac CMR sequences from a single frame by placing an ECG-free
nonlinear phase at the center of a latent flow-matching architecture. In this application, we choose the ED frame as the \emph{anchor frame} to generate the whole cine sequence.
A frozen segmentation network derives the cardiac phase $\phi_t$ at time t from the
per-frame LV area curve, encoding each frame as
$\mathbf{p}_t{=}[\sin\phi_t,\,\cos\phi_t]$, a compact, periodicity-aware
representation sensitive to the asymmetric contraction velocity that the linear
phase assumption misses.
A rectified flow model~\cite{liu2022rectified} then transports the ED anchor's
latent features $\mathbf{z}_0$ toward the full-sequence's latent features $\mathbf{z}_1$ along a
straight conditional trajectory, jointly conditioned on the cardiac phase and slice
position, with single-step inference supported natively by a
rectified-flow objective~\cite{sfv}.
The predicted latent velocity is decoded by a deformation decoder into a
diffeomorphic displacement field via scaling-and-squaring
integration~\cite{arsigny2006}, warping ED pixel intensities
\emph{directly}, bypassing the variational autoencoder (VAE)~\cite{kingma2013vae} decoding phase and preserving sharpness that
VAE reconstruction would otherwise blur.
Training is further guided by a \emph{VolumeCurveLoss} that differentiably
supervises the synthesized LV volume trajectory by applying the frozen
segmentation network exclusively to real CMR frames, providing
physiologically grounded gradient signal without inducing the domain-shift
artifacts that arise when segmentors encounter generated images.

Our main contributions are:
\begin{itemize}
    \item \textbf{ECG-free nonlinear phase estimation.}
        A segmentation-based LV area curve yields a physiologically
        interpretable cardiac phase signal capturing the asymmetric dynamics
        of systole and diastole, with no ECG hardware required at training
        or inference.
  \item \textbf{Phase-conditioned latent flow matching.}
        A rectified flow model conditioned on cardiac phase and slice position
        synthesizes the complete cardiac motion trajectory in latent space via
        single-step inference.
  \item \textbf{Deformation-decoded pixel synthesis.}
        A diffeomorphic deformation decoder converts the predicted latent
        velocity into a per-frame displacement that warps ED pixel intensities
        directly, eliminating VAE reconstruction blur.
\end{itemize}

\section{Related Work}

\paragraph{Cine CMR synthesis.}
Three paradigms have addressed cine synthesis.
Registration-based methods~\cite{dragnet, krebs2021} warp a reference frame via predicted displacement fields but treat frame index as a linear phase proxy, erasing systole--diastole velocity asymmetry and generalizing poorly across pathologies~\cite{regmagic}.
GAN and diffusion models~\cite{gancmri, heartbeat, echodiff_icassp, biomedvid} achieve visual realism but lack physiological supervision, leaving EF unconstrained; diffusion models further require tens to hundreds of denoising steps at inference, precluding real-time clinical deployment.
ECG-conditioned approaches~\cite{echopulse, ecgflowcmr} restore temporal coherence but require concurrent ECG acquisition, limiting applicability to retrospective or resource-limited datasets.

\paragraph{Flow matching for video generation.}
Rectified flow~\cite{liu2022rectified, lipman2022flow} enables single-step generation by
training along straight conditional trajectories in the latent space.
SF-V~\cite{sfv} extends this paradigm to videos, demonstrating that
single-forward-pass video synthesis is practical for high-resolution outputs.
EchoLVFM~\cite{echolv} applies latent flow matching to echocardiography,
the concurrent work most closely related to ours. However, it targets echocardiography (not CMR), omits a deformation decoder, and does not incorporate explicit volume-curve supervision.

\paragraph{Diffeomorphic deformation in medical imaging.}
Scaling-and-squaring integration~\cite{arsigny2006} provides a
computationally efficient route to diffeomorphic transformations.
VoxelMorph~\cite{voxelmorph} popularized learning-based diffeomorphic
registration for medical images.
These tools have been deployed almost exclusively in the \emph{registration}
paradigm, aligning a moving image to a fixed target.
In contrast, PhaseFlow repurposes this machinery as a \emph{generative decoder}: displacement is \emph{synthesized} from a latent velocity prediction rather than estimated between two observed images, replacing the VAE decoder while maintaining anatomical plausibility.

\section{Methods}
\begin{figure*}[!ht]
  \centering
  \includegraphics[width=\linewidth]{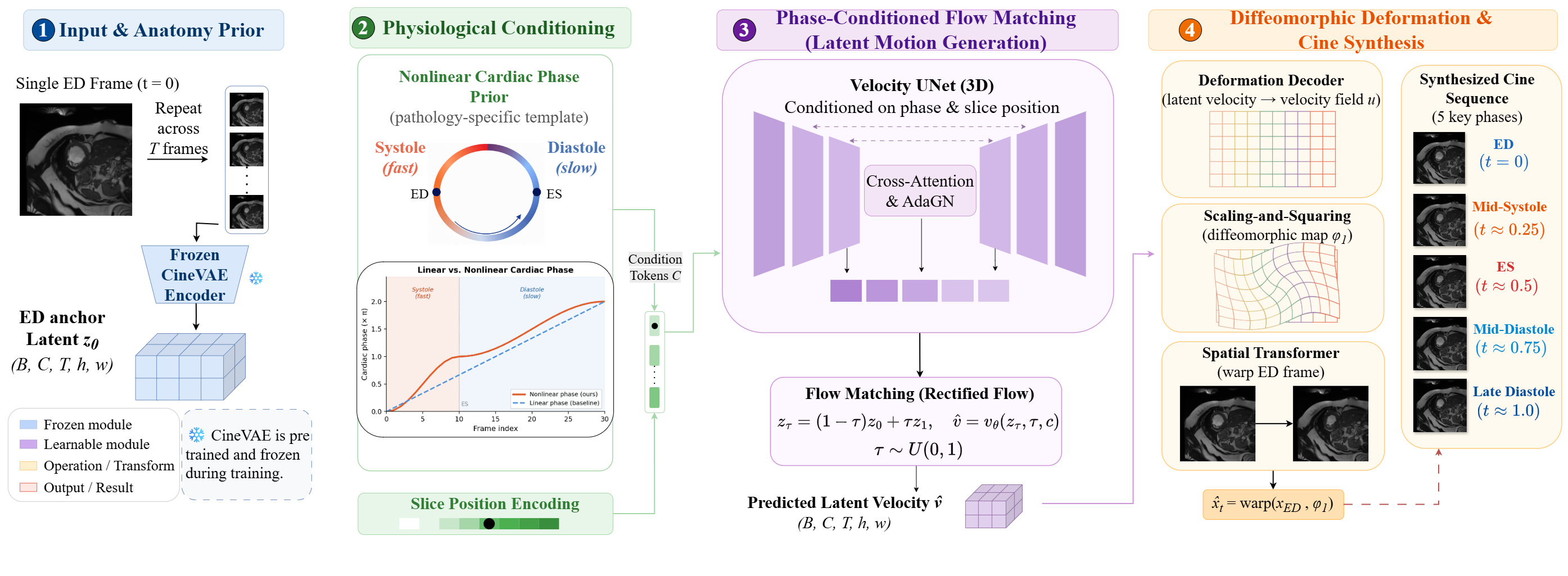}
    \caption{\textbf{PhaseFlow} encodes the ED frame into latent features, predicts a phase-conditioned latent velocity via rectified flow, and decodes a diffeomorphic displacement field that warps ED pixels into a fully synthesized cine sequence.}
  \label{fig:architecture}
\end{figure*}

Figure~\ref{fig:architecture} shows the overall architecture of the proposed PhaseFlow. Given the ED frame $\mathbf{x}_{\mathrm{ED}} \in \mathbb{R}^{H \times W}$
and a cardiac phase schedule $\{\phi_t\}_{t=1}^{T}$, PhaseFlow proceeds in
four stages:
(i)~a frozen VAE encoder maps the ED frame into the anchor latent $\mathbf{z}_0$;
(ii)~a nonlinear cardiac phase prior and slice position encoding form condition tokens $\mathbf{c}$;
(iii)~a phase-conditioned rectified flow model predicts the latent velocity conditioned on phase and slice position;
(iv)~a deformation decoder converts the latent velocity into a diffeomorphic displacement field that warps ED pixel intensities into a synthesized cine sequence.


\subsection{Problem Formulation}
\label{sec:problem}

Let $\mathbf{x}_{\mathrm{ED}} \in \mathbb{R}^{H \times W}$ denote the single ED CMR frame. The synthesized cine CMR sequence $\{\hat{\mathbf{x}}_1, \ldots, \hat{\mathbf{x}}_T\}$ (frame index $t \in \{1,\dots,T\}$) must be both \emph{perceptually realistic} and \emph{physiologically faithful}. This task is fundamentally ill-posed: cardiac functional assessment requires complete temporal sequences rather than isolated frames~\cite{jiang2024cmr}, and single-frame EF prediction is known to significantly underperform sequential models~\cite{echonet}. Successful synthesis therefore requires an external temporal signal, which we derive from the nonlinear cardiac phase (Sec.~\ref{sec:phase}).

\subsection{ECG-Free Nonlinear Phase Estimation}
\label{sec:phase}

We extract a physiologically grounded phase signal directly from the cine
CMR sequence without ECG dependency.
A frozen segmentation network (SegUNet)~\cite{ronneberger2015unet, isensee2021nnunet} segments each frame to obtain the
per-frame LV area $\{a_t\}_{t=1}^T$.
We define the cumulative area change as
\begin{equation}
  \phi_t = 2\pi \cdot \frac{\sum_{k=1}^{t} |a_k - a_{k-1}|}
                            {\sum_{k=1}^{T} |a_k - a_{k-1}|},
  \label{eq:phase}
\end{equation}
mapping the cardiac cycle onto $[0, 2\pi]$ proportionally to the
\emph{true dynamics of myocardial contraction} rather than a temporal frame index.
Each frame is then encoded as
$\mathbf{p}_t = [\sin\phi_t,\;\cos\phi_t] \in \mathbb{R}^2$,
a compact, periodicity-aware representation.

\begin{figure}[ht]
  \centering
  \includegraphics[width=\columnwidth]{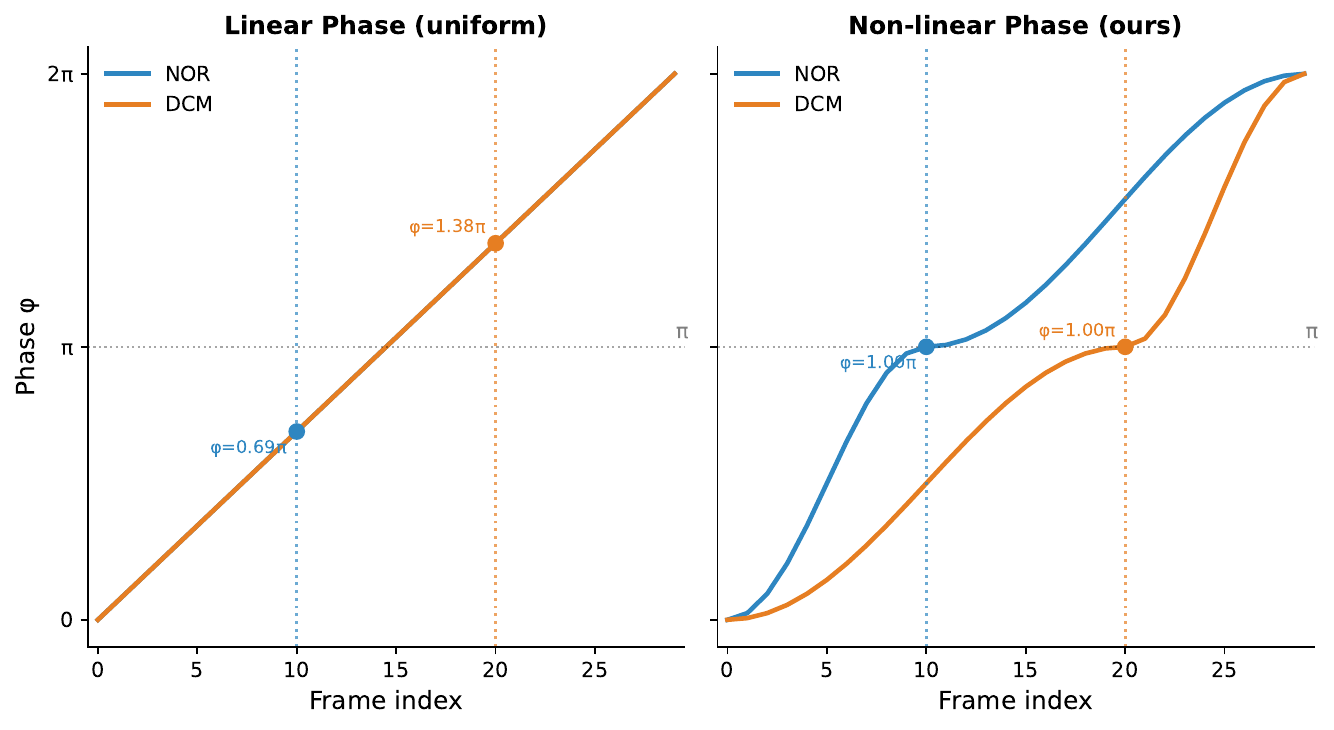}
  \caption{Linear phase (left) scatters ES across $\phi$ values; nonlinear phase (right) aligns ES at $\phi\!\approx\!\pi$ for different pathological types (e.g., normal and dilated cardiomyopathy).}
  \label{fig:toy_phase_b}
\end{figure}

\paragraph{Why nonlinear?}
Systole and diastole have markedly
different durations across patients and pathologies.
A linear phase $\phi_t^{\text{lin}} = 2\pi t / T$ uniformly spaces frames,
mapping ES to different $\phi$ values depending on heart rate, erasing
cross-patient consistency (Figure~\ref{fig:toy_phase_b}, left).
The proposed nonlinear phase instead accumulates $\phi_t$ proportionally to
the actual myocardial displacement $|a_k - a_{k-1}|$ such that ES naturally aligns
at $\phi \approx \pi$ across patients regardless of heart rate or pathology
(Figure~\ref{fig:toy_phase_b}, right).

\paragraph{Pathology-specific templates for ECG-free inference.}
During training, $\phi_t$ is computed per subject from the full cine sequence.
At inference, only the ED frame is available.
We pre-compute mean phase templates for each pathological group in the full dataset
by averaging the normalized phase curves across
training subjects within each group.
Given one pathology label (obtainable from clinical metadata or a simple
classifier), its corresponding template provides the phase schedule.
This design is motivated by evidence that the temporal evolution of LV
shape throughout the cardiac cycle is pathology-discriminative~\cite{zheng2019explainable},
implying that each pathology group exhibits a characteristic phase trajectory
distinct from others (visualized in Appendix~\ref{app:phase_templates}).

\subsection{Phase-Conditioned Latent Flow Matching}
\label{sec:flow}

We formulate cine synthesis as an optimal transport in a shared latent space
using rectified flow~\cite{liu2022rectified}.

\paragraph{Latent space construction.}
A pretrained, frozen three-dimensional (3D) VAE encoder~\cite{kingma2013vae, rombach2022ldm} $E_{\mathrm{vae}}$ maps the full cine
sequence $\mathbf{X} \in \mathbb{R}^{1 \times T \times H \times W}$
to the target latent $\mathbf{z}_1 = E_{\mathrm{vae}}(\mathbf{X})
\in \mathbb{R}^{C \times T \times h \times w}$ ($h = H/8$, $w = W/8$; temporal
dimension preserved). The anchor latent is built from the ED frame alone by
tiling $\mathbf{x}_{\mathrm{ED}}$ (Sec.~\ref{sec:problem}) $T$ times into a
\emph{temporally static sequence} $\mathrm{repeat}(\mathbf{x}_{\mathrm{ED}},\,T)$:
\begin{equation}
  \mathbf{z}_0 = E_{\mathrm{vae}}\bigl(\mathrm{repeat}(\mathbf{x}_{\mathrm{ED}},\, T)\bigr) + \alpha\,\boldsymbol{\varepsilon},
  \quad \boldsymbol{\varepsilon} \sim \mathcal{N}(\mathbf{0},\,\mathbf{I}).
  \label{eq:z0}
\end{equation}
Because $E_{\mathrm{vae}}$ uses 3D temporal kernels, this static input is
resolved as carrying zero motion rather than as a replicated single-frame
latent: $\mathbf{z}_0$ thus encodes a hypothetical \emph{arrested-heart} state,
and flow matching learns transport from it to the true dynamics $\mathbf{z}_1$.
The perturbation $\alpha\boldsymbol{\varepsilon}$ (with $\alpha$ analyzed in
Appendix~\ref{app:latent_scale}) expands this deterministic start into a
neighborhood, keeping the learned velocity field stable when $\mathbf{z}_0$ is
built from a single real ED frame at inference.

\paragraph{Rectified flow.}
Given a uniformly distributed flow time $\tau \sim \mathcal{U}(0, 1)$, the
interpolated latent feature at $\tau$ is
$\mathbf{z}_\tau = (1 - \tau)\,\mathbf{z}_0 + \tau\,\mathbf{z}_1$.
Differentiating with respect to $\tau$ gives the ground-truth \emph{feature velocity}
$\mathbf{v}^* = {d\mathbf{z}_\tau}/{d\tau} = \mathbf{z}_1 - \mathbf{z}_0$,
which turns out to be \emph{constant} despite $
\tau$.
A 3D UNet~\cite{ronneberger2015unet} $v_\theta$ (VelocityUNet applied on the spatiotemporal
dimensions) is trained to predict
this velocity, conditioned on the flow time $\tau$ and a conditioning vector $\mathbf{c}$:
\begin{equation}
  \hat{\mathbf{v}}(\tau) = v_\theta(\mathbf{z}_\tau,\;\tau,\;\mathbf{c}),
  \qquad
  \mathbf{c} = \bigl[\mathbf{e}_{\mathrm{phase}},\;\mathbf{e}_{\mathrm{slice}}\bigr],
  \label{eq:velocity}
\end{equation}
where $\mathbf{e}_{\mathrm{phase}} \in \mathbb{R}^{T \times d}$ is our proposed
per-frame phase embedding produced by a small multilayer perceptron (MLP; the PhaseProjector) from
$\mathbf{p}_t$, and
$\mathbf{e}_{\mathrm{slice}} \in \mathbb{R}^{d}$ encodes the spatial slice
position.
To enable cross-frame reasoning, each resolution level of $v_\theta$ incorporates
a temporal attention block that exchanges information across all $T$ frames
simultaneously, allowing the model to synthesize temporally coherent motion
rather than processing each frame independently.
The per-frame phase embedding $\mathbf{e}_{\mathrm{phase}}$ is injected at every resolution level via cross-attention~\cite{vaswani2017attention}, enabling spatially adaptive phase conditioning.
At inference, only $\mathbf{z}_0$ is available. Since $\mathbf{v}^*$ is
constant along the straight-line path, the velocity network can be queried
at any $\tau$ with $\mathbf{z}_0$ as input; we use $\tau=1$, evaluating the velocity network at the trajectory terminus,
aligned with the phase-conditioned target frame:
$\hat{\mathbf{v}}(1) = v_\theta(\mathbf{z}_0,\,1,\,\mathbf{c})$,
requiring no iterative ordinary differential equation (ODE) integration~\cite{liu2022rectified, sfv}.
The predicted latent velocity $\hat{\mathbf{v}}(1)$ is then decoded into
the image space by a deformation decoder (Sec.~\ref{sec:deform}).

\subsection{Deformation-Decoded Pixel Synthesis}
\label{sec:deform}

Rather than decoding through a VAE decoder (which would introduce reconstruction blur),
we decode the predicted latent velocity $\hat{\mathbf{v}}(1)$ into a diffeomorphic
displacement field and use it to directly warp the original ED frame intensities.

\paragraph{Deformation decoder.}
A transposed-convolutional decoder $D_{\mathrm{deform}}$ maps the predicted
latent velocity $\hat{\mathbf{v}}(1) \in \mathbb{R}^{C \times T \times h \times w}$
to a continuous velocity field
$\mathbf{u} \in \mathbb{R}^{2 \times T \times H \times W}$ in the image space.
The decoder acts frame-wise: each of the $T$ temporal slices is processed
independently by a stack of two-dimensional (2D) transposed convolutions that upsamples the
spatial resolution $8\times$ ($h \times w \rightarrow H \times W$) while
projecting the $C$ latent channels onto the two displacement components
(horizontal and vertical). The temporal axis is carried through unchanged,
as temporal coherence is already established in latent space by $v_\theta$.

\paragraph{Scaling-and-squaring integration.}
Following~\cite{arsigny2006}, we obtain the diffeomorphic displacement
$\boldsymbol{\varphi}_t$ via $K$ composition steps:
\begin{equation}
  \boldsymbol{\varphi}_t = \underbrace{
    \bigl(\tfrac{\mathbf{u}}{2^K} \circ \cdots \circ \tfrac{\mathbf{u}}{2^K}\bigr)
  }_{2^K \text{ compositions}}.
  \label{eq:ss}
\end{equation}
It guarantees $\det J_{\boldsymbol{\varphi}_t}(\mathbf{r}) > 0$ at every pixel
location $\mathbf{r}$, where $J_{\boldsymbol{\varphi}_t}(\mathbf{r}) =
\partial \boldsymbol{\varphi}_t / \partial \mathbf{r}$ is the spatial Jacobian of
the deformation field
(empirically verified in Appendix~\ref{app:jacobian}).
Each synthesized frame is
\begin{equation}
  \hat{\mathbf{x}}_t = \mathrm{warp}\bigl(\mathbf{x}_{\mathrm{ED}},\;
    \boldsymbol{\varphi}_t\bigr),
  \label{eq:warp}
\end{equation}
where $\mathrm{warp}$ is performed with bilinear grid sampling~\cite{jaderberg2015stn}.

\subsection{Training Objectives}
\label{sec:loss}

Training combines four objectives, assembled in Eq.~\eqref{eq:total_loss}.
In every term, the flow time $\tau \sim \mathcal{U}(0,1)$ and anchor noise
$\boldsymbol{\varepsilon} \sim \mathcal{N}(\mathbf{0},\mathbf{I})$ are redrawn
each step and thus enter as an expectation, while the frame index
$t \in \{1,\dots,T\}$ enters as an average over the $T$ frames.

\paragraph{Flow matching loss (core).}
The standard rectified flow matching objective, evaluated in latent space:
\begin{equation}
  \mathcal{L}_{\mathrm{flow}} = \mathbb{E}_{\tau,\,\boldsymbol{\varepsilon}}
  \bigl[\lVert v_\theta(\mathbf{z}_\tau,\;\tau,\;\mathbf{c}) -
  (\mathbf{z}_1 - \mathbf{z}_0) \rVert^2\bigr].
  \label{eq:lflow}
\end{equation}

\paragraph{Local normalized cross-correlation.}
An image-space registration loss between each warped ED frame and the
corresponding ground-truth frame $\mathbf{x}_t^{\mathrm{GT}}$, robust to
intensity offsets:
\begin{equation}
  \mathcal{L}_{\mathrm{lncc}} = 1 - \frac{1}{T} \sum_{t=1}^{T} \mathrm{LNCC}\bigl(
    \mathrm{warp}(\mathbf{x}_{\mathrm{ED}},\,\boldsymbol{\varphi}_t),\;
    \mathbf{x}_t^{\mathrm{GT}}\bigr).
  \label{eq:llncc}
\end{equation}

\paragraph{Bending energy.}
A smoothness regularizer~\cite{rueckert1999nonrigid} penalizing second-order spatial derivatives of each
per-frame displacement field:
\begin{equation}
  \begin{aligned}
  \mathcal{L}_{\mathrm{bend}} = \frac{1}{T} \sum_{t=1}^{T} \Bigl(
  &\lVert \partial^2 \boldsymbol{\varphi}_t / \partial x^2 \rVert^2 \\
  &+ \lVert \partial^2 \boldsymbol{\varphi}_t / \partial y^2 \rVert^2
  + 2\lVert \partial^2 \boldsymbol{\varphi}_t / \partial x\,\partial y \rVert^2
  \Bigr).
  \end{aligned}
  \label{eq:lbend}
\end{equation}

\paragraph{VolumeCurveLoss.}
A differentiable objective that supervises the synthesized LV volume
trajectory without domain shift.
The frozen SegUNet processes only real CMR frames: it segments the ED frame
to obtain a soft LV mask $\mathbf{m}_{\mathrm{ED}}$ and each ground-truth
frame to obtain the target area $a_t^{\mathrm{GT}}$; generated images are
\emph{never} passed to the segmentation network:
\begin{equation}
  \mathcal{L}_{\mathrm{vol}} = \frac{1}{T} \sum_{t=1}^{T}
  \frac{\bigl|\,\textstyle\sum_{\mathbf{r} \in \Omega}
    \mathrm{warp}(\mathbf{m}_{\mathrm{ED}},\,\boldsymbol{\varphi}_t)(\mathbf{r})
    - a_t^{\mathrm{GT}}\bigr|}{\mathrm{EDV}}.
  \label{eq:lvol}
\end{equation}
Here $\mathbf{r}$ indexes spatial pixel locations over the image domain
$\Omega$, so $\sum_{\mathbf{r} \in \Omega}$ counts the predicted LV area at
frame $t$, and $\mathrm{EDV}$ is the ED volume used for normalization.
The loss is fully differentiable through the bilinear warp back to VelocityUNet.
By confining SegUNet to real MRI frames, domain shift is avoided and gradients remain reliable.

\paragraph{Total objective.}
The four terms are combined as
\begin{equation}
  \mathcal{L} = \lambda_{\mathrm{flow}}\,\mathcal{L}_{\mathrm{flow}}
  + \lambda_{\mathrm{lncc}}\,\mathcal{L}_{\mathrm{lncc}}
  + \lambda_{\mathrm{bend}}\,\mathcal{L}_{\mathrm{bend}}
  + \lambda_{\mathrm{vol}}\,\mathcal{L}_{\mathrm{vol}},
  \label{eq:total_loss}
\end{equation}
with weights $\lambda$ balancing the latent-space transport objective against
the three image-space terms; their values are reported in the implementation
details (Sec.~\ref{sec:setup}).

\section{Experiments and Results}

\subsection{Experiment Setup}
\label{sec:setup}

\paragraph{Dataset.}
We used the ACDC cardiac MRI benchmark dataset for evaluation~\cite{bernard2018acdc}, which contains 150 subjects from five pathological groups: normal control (NOR), dilated cardiomyopathy (DCM), hypertrophic cardiomyopathy (HCM), prior myocardial infarction (MINF), and abnormal right ventricle (ARV), with 30 subjects per group. Each subject includes a cine CMR sequence with ED/ES annotation. We split the dataset into 105 training subjects, 15 validation subjects, and 30 test subjects in a stratified manner  with the five pathological groups evenly distributed across all splits.
To assess cross-dataset generalization, we additionally evaluate on the Multi-Centre, Multi-Vendor \& Multi-Disease (M\&Ms) dataset~\cite{campello2021mms,martinisla2023mms2}, which comprises 136 test subjects acquired across four scanner vendors and multiple clinical centres. No M\&Ms data was used during training; the model was applied zero-shot using ACDC pathology-specific phase templates.

\paragraph{Implementation.}
The SegUNet is pre-trained on ACDC images with segmentation annotations
and kept frozen throughout flow-matching training.
The frozen VAE (3D, spatial compression $8\times$, temporal preserved)
was pretrained on ACDC for 100 epochs.
The flow-matching network (VelocityUNet) and deformation decoder were trained jointly with AdamW~\cite{loshchilov2018adamw} (lr = $10^{-4}$),
batch size 4, converging within $\sim$165 epochs ($\sim$5.5 hours) on a single NVIDIA RTX A6000 (48\,GB) GPU.
Noise scale $\alpha = 0.1$; scaling-and-squaring steps $K = 7$.
Loss weights (Eq.~\eqref{eq:total_loss}) were set to
$\lambda_{\mathrm{flow}} = 1.0$, $\lambda_{\mathrm{lncc}} = 1.0$,
$\lambda_{\mathrm{bend}} = 0.001$, and $\lambda_{\mathrm{vol}} = 2.0$.
At inference, pathology-specific phase templates were used
(Sec.~\ref{sec:phase}; see Appendix~\ref{app:implementation} for full architecture details).

\paragraph{Evaluation metrics.}
We organized evaluation into two categories:
\begin{itemize}
  \item \textbf{Image quality:} PSNR~(pixel-level fidelity), SSIM~\cite{wang2004ssim}~(structural similarity), LPIPS~\cite{zhang2018lpips}~(perceptual distance), FID~\cite{heusel2017fid}~(generative quality).
  \item \textbf{Physiological fidelity:} EF~MAE~(ejection fraction mean absolute error),
    Vol~Corr~(volume curve Pearson correlation), Vol~$R^2$~(volume curve
    coefficient of determination), Vol~MAE~(normalized per-frame volume
    error).
\end{itemize}
Physiological metrics were computed by warping the ground-truth ED
segmentation mask through the predicted displacement field, avoiding
segmentation domain shift in the evaluation pipeline.
Formal definitions and the motivation for each metric are given in
Appendix~\ref{app:metrics}.

\paragraph{Baselines.}
We compared against five methods spanning different paradigms:
\textbf{(1)~ED~Repeat}: repeating the ED frame $T$ times (trivial lower
bound);
\textbf{(2)~ConvLSTM}~\cite{shi2015convlstm}: a recurrent spatiotemporal prediction model;
\textbf{(3)~Direct~Registration}: a VAE encoder followed by latent
regression to per-frame displacement fields (no flow matching);
\textbf{(4)~CVAE}: a variational autoencoder that generates frames conditioned on the ED frame;
and \textbf{(5)~EchoDiff}: the conditional video diffusion model of~\citet{echodiff_icassp}, adapted to CMR by replacing echocardiography frames with ACDC cine slices and training for 100k steps.
We further exclude four otherwise-natural baselines (see Appendix~\ref{app:baselines}): DragNet~\cite{dragnet} (autoregressive drift breaks physiological evaluation), registration methods such as VoxelMorph~\cite{voxelmorph} (require a target frame at inference), ECG-conditioned approaches~\cite{echopulse, ecgflowcmr} (violate our ECG-free setting), and EchoLVFM~\cite{echolv} (no public implementation).
All baselines were retrained on the same ACDC split under identical conditions.

\subsection{Comparison to Baselines}

\begin{table*}[htbp]
\centering
\small
\setlength{\tabcolsep}{3.5pt}
\begin{tabular}{l cc cc cc cc}
\toprule
& \multicolumn{4}{c}{\textbf{Image Quality}} &
  \multicolumn{4}{c}{\textbf{Physiological Fidelity}} \\
\cmidrule(lr){2-5} \cmidrule(lr){6-9}
Method
  & PSNR$\uparrow$ & SSIM$\uparrow$ & LPIPS$\downarrow$ & FID$\downarrow$
  & \makebox[1.6cm]{EF MAE$\downarrow$} & Vol Corr$\uparrow$ & Vol $R^2$$\uparrow$ & Vol MAE$\downarrow$ \\
\midrule
ED Repeat
  & 30.63 & 0.950 & \textbf{0.022} & 18.22
  & 51.37 & --- & $-$2.050 & 0.245 \\
ConvLSTM
  & 27.57 & 0.903 & 0.137 & 79.94
  & 48.12 & 0.082 & $-$1.539 & 0.230 \\
Direct Reg
  & \textbf{31.89} & \underline{0.955} & 0.031 & \underline{17.26}
  & 19.66 & \textbf{0.924} & $-$0.105 & \underline{0.116} \\
CVAE
  & 30.89 & 0.949 & 0.037 & 26.04
  & 22.29 & \underline{0.921} & \underline{$-$0.038} & 0.122 \\
EchoDiff~\citep{echodiff_icassp}
  & 15.20 & 0.219 & 0.391 & 115.17
  & \textbf{15.47} & 0.414 & $-$1.000 & 0.211 \\
\midrule
\textbf{PhaseFlow}$^\dagger$
  & \underline{31.72} & \textbf{0.956} & \underline{0.023} & \textbf{12.72}
  & \underline{17.79} & \ 0.867 & \textbf{0.363} & \textbf{0.101} \\
\bottomrule
\end{tabular}
\caption{\textbf{Main results on ACDC test set} (30 patients).
Best in \textbf{bold}, second-best \underline{underlined}.}
\label{tab:main}
\end{table*}

Table~\ref{tab:main} reports the comparison against the five baselines on the
ACDC test set, grouped into the two metric families of image quality (PSNR,
SSIM, LPIPS, FID) and physiological fidelity (EF~MAE, Vol~Corr, Vol~$R^2$,
Vol~MAE). Two observations stand out:

\paragraph{(1)~PhaseFlow achieved the best physiological fidelity.}
PhaseFlow was the \emph{only} method achieving a positive Vol~$R^2$ ($0.363$), meaning it was the only model whose predicted LV volume curve fit the ground truth better than a constant-mean predictor; all other methods yielded negative $R^2$, indicating systematic failure to capture ventricular contraction magnitude.
EchoDiff recorded the numerically lowest EF~MAE ($15.47\%$), but at the cost of severely degraded image quality (SSIM $0.219$, FID $115.17$) and a Vol~$R^2$ of $-1.00$, suggesting it incidentally reproduced ED- and ES-sized blurry frames rather than faithfully tracking the volume trajectory.
PhaseFlow achieved the second-best EF~MAE ($17.79\%$) with fully preserved image quality, surpassing Direct Registration ($19.66\%$) which directly optimised pixel-level losses.
This confirmed that the combination of nonlinear phase conditioning and VolumeCurveLoss provided physiological grounding that pixel-level objectives alone could not.

\paragraph{(2)~PhaseFlow achieved strong image quality.}
PhaseFlow achieved the best SSIM ($0.956$) and FID ($12.72$), with PSNR
($31.72$~dB) comparable to Direct Registration ($31.89$~dB, $0.955$) and
substantially outperforming the recurrent baseline (ConvLSTM: $27.57$~dB).
This was notable because a model optimizing pixel-level losses (Direct Reg)
achieved stronger PSNR but higher FID, reflecting the
well-known perception--distortion trade-off~\cite{blau2018perception}:
minimizing point-wise reconstruction error yielded blurry,
distribution-atypical outputs.
PhaseFlow's deformation decoder sidestepped this trade-off by sourcing all
pixel values from the real ED frame.

\subsection{Ablation Study}
\label{sec:ablation}

\begin{table*}[htbp]
\centering
\small
\setlength{\tabcolsep}{3.5pt}
\begin{tabular}{l cccc cccc}
\toprule
& \multicolumn{4}{c}{\textbf{Image Quality}} &
  \multicolumn{4}{c}{\textbf{Physiological Fidelity}} \\
\cmidrule(lr){2-5} \cmidrule(lr){6-9}
Variant
  & PSNR$\uparrow$ & SSIM$\uparrow$ & LPIPS$\downarrow$ & FID$\downarrow$
  & EF MAE$\downarrow$ & Vol Corr$\uparrow$ & Vol $R^2$$\uparrow$ & Vol MAE$\downarrow$ \\
\midrule
Full (pathology phase)
  & 31.72 & 0.956 & 0.023 & 12.72
  & 17.79 & 0.867 & 0.363 & 0.101 \\
\midrule
A1: VAE decoder
  & 27.28 & 0.882 & 0.128 & 89.63
  & 20.22 & 0.888 & $-$0.548 & 0.128 \\
A2: Linear phase
  & 31.48 & 0.954 & 0.024 & 12.91
  & 21.30 & 0.827 & 0.069 & 0.114 \\
A3: No LNCC 
  & 15.17 & 0.238 & 0.176 & 50.25
  & 14.88 & 0.876 & 0.295 & 0.094 \\
A4: No vol loss 
  & 31.76 & 0.955 & 0.023 & 11.43
  & 26.80 & 0.869 & 0.071 & 0.133 \\
\bottomrule
\end{tabular}
\caption{\textbf{Ablation study} on the ACDC test set.
Each row removes one component from the full PhaseFlow model.}
\label{tab:ablation}
\end{table*}

Table~\ref{tab:ablation} shows ablation of four key components.
Each removal produced a distinct failure mode, confirming that all components
addressed complementary aspects of the problem:

\paragraph{A1: VAE decoder.}
Replacing the deformation decoder with a standard VAE decoder caused a large
drop in image quality (PSNR $27.28$~dB vs.\ $31.72$~dB, FID $89.63$
vs.\ $12.72$) and degraded physiological fidelity (Vol~$R^2$ dropped from
$0.363$ to $-0.548$).
The ${\sim}4.4$~dB PSNR gap and the order-of-magnitude FID increase directly
demonstrate the pixel-preservation advantage of deformation decoding: by
sourcing all pixel values from the real ED frame rather than reconstructing
them through a bottleneck, the deformation decoder avoids VAE reconstruction
blur while simultaneously producing more distribution-faithful outputs.

\paragraph{A2: Linear phase.}
Replacing the nonlinear phase with uniform $\phi_t^{\text{lin}} = 2\pi t / T$
degraded EF~MAE from $17.79\%$ to $21.30\%$ and Vol~$R^2$ from $0.363$ to
$0.069$.
The effect was incremental but consistent: nonlinear phase provided a more
faithful temporal parameterization, particularly for pathologies with
pronounced systole--diastole asymmetry.

\paragraph{A3: No LNCC loss.}
Removing the image-space registration loss caused a \emph{major failure}:
PSNR dropped to $15.17$~dB and SSIM dropped to $0.238$.
Without direct pixel-level supervision, the deformation decoder had no
signal to learn meaningful spatial displacements; the flow matching loss in
latent space alone was insufficient to guide image-space warp quality.
Interestingly, physiological metrics remained reasonable,
suggesting the VolumeCurveLoss provided certain implicit spatial guidance, but
the images were still visually unusable.

\paragraph{A4: No volume loss.}
Removing VolumeCurveLoss preserved image quality (PSNR $31.76$~dB) but
severely degraded physiological fidelity: EF~MAE rose from $17.79\%$ to
$26.80\%$ and Vol~$R^2$ dropped from $0.363$ to $0.071$.
This was the clearest evidence that pixel-level losses alone could not teach the
model how the heart contracted; explicit volume-curve supervision
was indispensable for physiologically faithful synthesis.

\subsection{Zero-Shot Cross-Dataset Generalization}
\label{sec:mms}

\begin{figure*}[!t]
  \centering
  \includegraphics[width=0.95\linewidth]{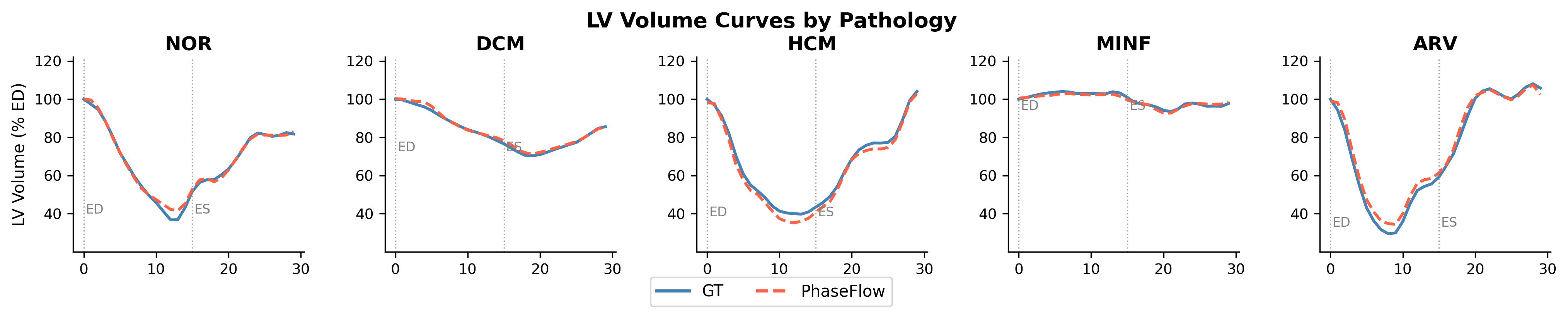}
  \caption{\textbf{LV volume curves.} Ground-truth (solid) vs.\ predicted (dashed) trajectories per pathology group.}
  \label{fig:volume_curves}
\end{figure*}

\begin{figure*}[!t]
  \centering
  \includegraphics[width=\linewidth]{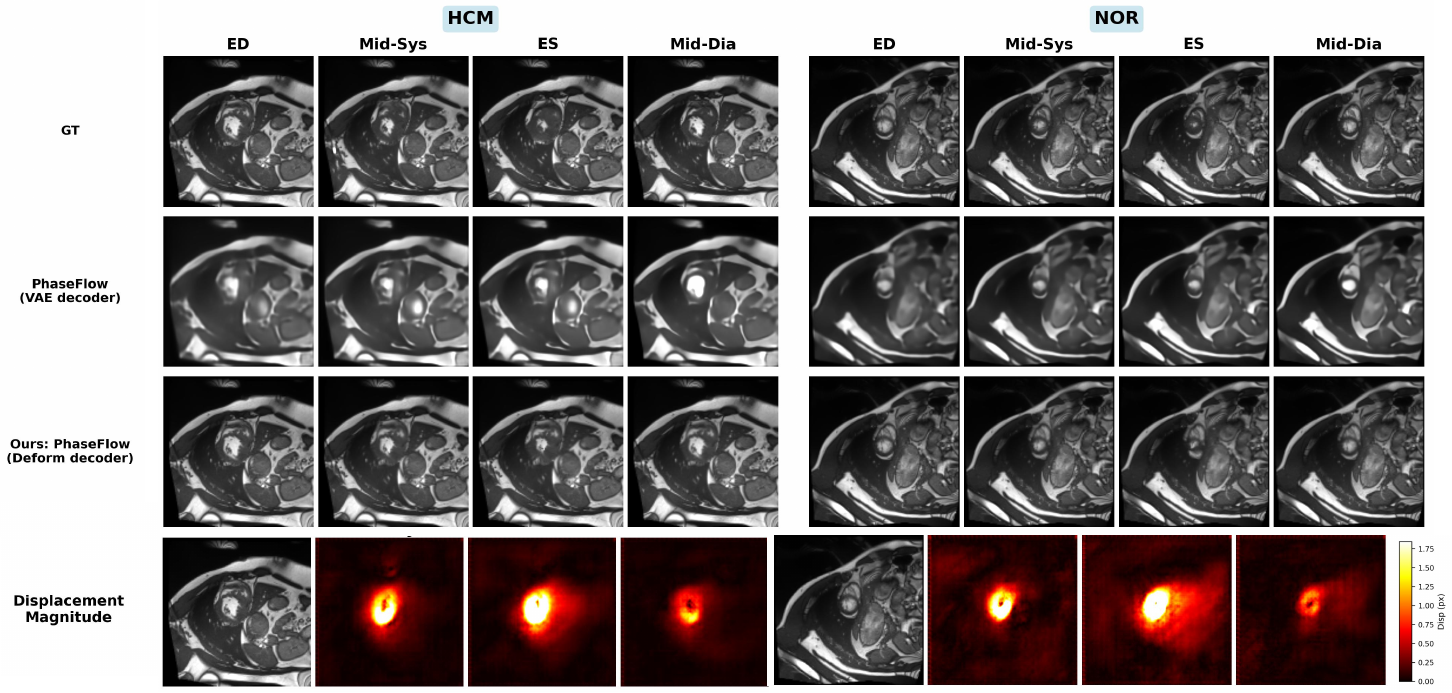}
  \caption{\textbf{Qualitative comparison} (HCM and NOR). Rows 1--3: GT, PhaseFlow-VAE, and our PhaseFlow (Deform) at ED, mid-systole, ES, and mid-diastole. Row 4: predicted displacement magnitude. PhaseFlow produces sharp, anatomically coherent frames; PhaseFlow-VAE shows characteristic reconstruction blur.}
  \label{fig:qualitative}
\end{figure*}

To test whether PhaseFlow could generalize beyond its training domain, we evaluated
the ACDC-trained model directly on the M\&Ms test set (136 patients, 1{,}621
slices) without any fine-tuning.
ACDC pathology-specific phase templates were applied at inference.

\begin{table}[htbp]
\centering
\setlength{\tabcolsep}{3pt}
\resizebox{\columnwidth}{!}{%
\begin{tabular}{l cc cc cc cc}
\toprule
& \multicolumn{4}{c}{\textbf{Image Quality}} &
  \multicolumn{4}{c}{\textbf{Physiological Fidelity}} \\
\cmidrule(lr){2-5} \cmidrule(lr){6-9}
Setting
  & PSNR$\uparrow$ & SSIM$\uparrow$ & LPIPS$\downarrow$ & FID$\downarrow$
  & EF MAE$\downarrow$ & Vol Corr$\uparrow$ & Vol $R^2$$\uparrow$ & Vol MAE$\downarrow$ \\
\midrule
ACDC \emph{(in-domain)}
  & 31.72 & 0.956 & 0.023 & 12.72
  & 17.79 & 0.867 & 0.363 & 0.101 \\
M\&Ms \emph{(zero-shot)}
  & 32.80 & 0.937 & 0.016 & \phantom{0}4.82
  & 19.89 & 0.845 & 0.309 & 0.114 \\
\bottomrule
\end{tabular}%
}
\caption{\textbf{In-domain vs.\ zero-shot generalization.}
PhaseFlow trained on ACDC and evaluated on M\&Ms without any adaptation.}
\label{tab:mms}
\end{table}

As shown in Table~\ref{tab:mms}, PhaseFlow transfers robustly to the unseen
domain: image quality even improves on the sharper, more anatomically diverse
M\&Ms acquisitions, while physiological fidelity remains close to in-domain
levels.
These results demonstrated that PhaseFlow's phase-conditioned motion priors generalized across scanner vendors and clinical sites without retraining, despite ACDC templates not perfectly representing M\&Ms acquisition dynamics.

\subsection{Qualitative Results}

Figure~\ref{fig:volume_curves} compares predicted and ground-truth LV volume
curves.
PhaseFlow captured the pathology-specific contraction patterns: normal hearts
showed the expected ${\sim}55{-}70\%$ EF, while DCM patients exhibited reduced
contraction and HCM patients showed preserved but hypertrophic dynamics.
The temporal alignment between predicted and ground-truth curves, driven by
the nonlinear phase conditioning, correctly reproduced the asymmetric
systole/diastole velocity profile.
Per-pathology qualitative results across all five ACDC groups are provided in Appendix~\ref{app:qualitative}.

Figure~\ref{fig:qualitative} shows synthesized frames and predicted displacement fields for representative HCM and NOR patients.
PhaseFlow produced sharp, anatomically coherent frames across the full cardiac cycle, with clearly visible myocardial thickening during systole and chamber expansion during diastole. In contrast, PhaseFlow-VAE exhibited the characteristic blur of VAE reconstruction, particularly at fine anatomical boundaries.
The displacement field visualizations showed smooth, anatomically plausible deformation concentrated around the myocardium.

\section{Conclusion}

We present PhaseFlow, a generative framework for synthesizing physiologically faithful cardiac cine MRI from a single end-diastolic frame without ECG dependency.
PhaseFlow addresses a key limitation of existing approaches, which either depend on explicit ECG acquisition to parameterize cardiac motion, or, being ECG-free but lacking physiological constraints, produce synthesized motion that drifts from true cardiac dynamics.
Instead, PhaseFlow conditions a flow-matching generator on a nonlinear cardiac phase supplied by a pathology-specific template. From this conditioned flow, a diffeomorphic decoder synthesizes a temporally coherent, anatomically consistent motion trajectory.
Extensive experiments on ACDC, together with zero-shot evaluation across scanner vendors and clinical sites, demonstrate the superior physiological fidelity and image realism of PhaseFlow.

\clearpage
\bibliography{refs}

\clearpage
\appendix

\section{Limitations and Broader Impact}
\label{app:limitations}

\paragraph{Limitations.}
The current EF~MAE of $17.79\%$ remains above the clinical diagnostic threshold of ${\sim}10\%$, and further validation on larger and more diverse cine CMR cohorts is needed before clinical deployment.
Evaluation is presently limited to CMR; extension to other spatiotemporal acquisitions such as echocardiography and lung MRI remains future work.
Additionally, phase estimation relies on SegUNet LV segmentation, so low image quality or heavy motion artifacts may degrade the phase signal and, consequently, the physiological fidelity of the synthesized sequence.

\paragraph{Broader impact.}
More broadly, PhaseFlow enables physiologically grounded cine sequence generation for retrospective cohort studies, resource-limited settings without ECG-gated infrastructure, and as a data augmentation strategy for downstream segmentation and disease classification tasks.

\section{Nonlinear Phase Templates: Empirical Analysis}
\label{app:phase_templates}

\subsection{Phase Template Visualization}
\label{app:phase_templates_b1}

\begin{figure}[htbp]
  \centering
  \includegraphics[width=\linewidth]{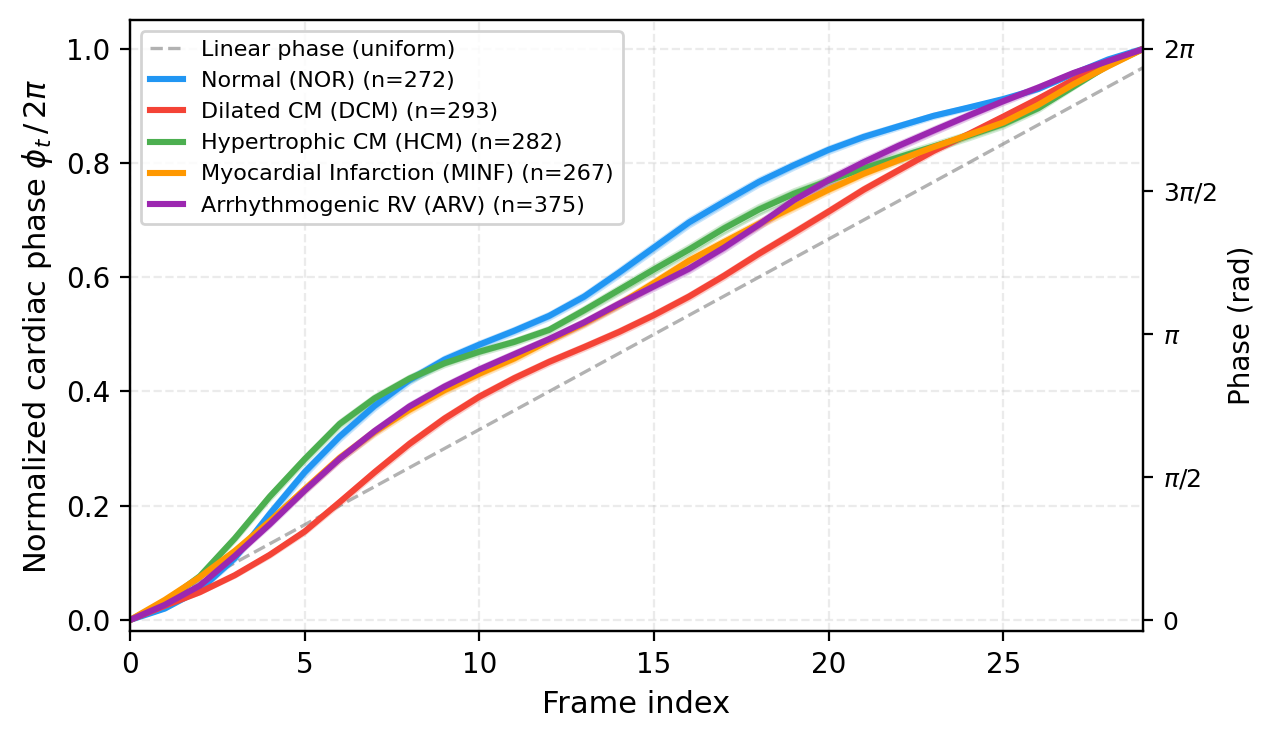}
  \caption{Per-pathology nonlinear phase templates (coloured) vs.\ linear baseline (grey dashed).
    All groups show a fast-then-slow profile: rapid systolic rise, then slower diastolic progression.}
  \label{fig:phase_templates}
\end{figure}

Figure~\ref{fig:phase_templates} visualises the per-group mean phase templates
$\{\bar{\phi}_t^{(g)}\}$ used at inference, computed by averaging the
nonlinear phase curves of all training subjects within each ACDC pathology
group.
The grey dashed line shows the uniform linear phase baseline.

All five pathology groups exhibit a consistent \emph{fast-then-slow} profile:
phase accumulates rapidly during the systolic contraction phase and more
gradually during diastolic relaxation, in contrast to the constant-rate linear
baseline.
Inter-group differences are visible: NOR subjects show earlier phase
progression, reflecting faster myocardial contraction, while DCM and MINF
subjects exhibit a delayed acceleration consistent with impaired contractility.

\FloatBarrier
\subsection{Intra-Group Consistency}
\label{app:phase_templates_b2}

To quantify how well a group-level template approximates individual patient
phase curves, we measure the per-frame root mean squared error (RMSE) of each
per-slice nonlinear phase curve relative to two candidate templates:
(i)~the \emph{nonlinear group mean} $\bar{\phi}_t^{(g)}$ (our method), and
(ii)~the \emph{linear phase} $\phi_t^{\text{lin}} = t/T$ (the baseline).

Because the group mean is computed from the same population, its RMSE equals
the intra-group standard deviation (zero bias by construction).
The linear phase introduces an additional systematic bias (the gap between
the nonlinear group mean and the diagonal visible in Figure~\ref{fig:phase_templates}), so
its RMSE satisfies $\text{RMSE}_{\text{lin}} = \sqrt{\text{bias}^2 + \sigma^2}
\geq \text{RMSE}_{\text{nl}}$.

\begin{figure*}[htbp]
  \centering
  \includegraphics[width=\linewidth]{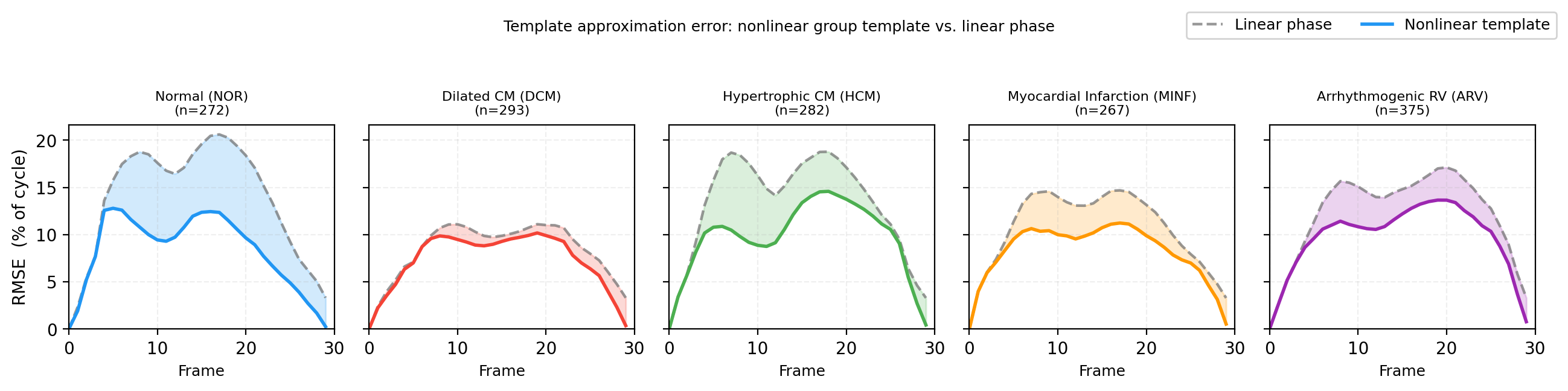}
  \caption{Per-frame phase RMSE: nonlinear group template (solid) vs.\ linear baseline (dashed).
    Shaded region: error gap. Nonlinear template reduces approximation error by up to $1.7\times$.}
  \label{fig:intra_group_consistency}
\end{figure*}

Figure~\ref{fig:intra_group_consistency} shows both RMSE curves for each
pathology group.
Across all groups, the nonlinear group template achieves a mean RMSE of
$7.3\text{--}9.7\%$ of the full cardiac cycle, while the linear phase incurs
$8.3\text{--}13.7\%$, an \textbf{up-to-$1.7\times$ reduction in approximation
error} (NOR group).
The shaded region between the two curves represents the error saved by using
the nonlinear group template instead of linear phase.
This demonstrates that the proposed phase representation substantially reduces
the uncertainty introduced by replacing per-patient ECG with a group-level
template at inference time.

\section{Diffeomorphism Verification}
\label{app:jacobian}

A central architectural claim of PhaseFlow is that the
\emph{Scaling-and-Squaring} (S\&S) integration of the velocity field
guarantees a diffeomorphic displacement field, i.e.\ a topology-preserving,
fold-free deformation with $\det(\mathbf{J}_\varphi)>0$ everywhere.
This appendix provides an empirical verification of that claim on the ACDC
test set.

\paragraph{Setup.}
The deformation decoder produces a 2-D velocity field
$\mathbf{u} \colon \mathbb{R}^{2}\!\to\!\mathbb{R}^{2}$, which is integrated
via $K=7$ S\&S squaring steps to yield the displacement field
$\varphi = \mathrm{exp}(\mathbf{u})$.
For a deformation $\varphi(\mathbf{x}) = \mathbf{x} + \mathbf{u}(\mathbf{x})$,
the Jacobian matrix and its determinant are
\[
\begin{aligned}
  \mathbf{J}_\varphi
  &=
  \begin{pmatrix}
    1+\tfrac{\partial u_x}{\partial x} & \tfrac{\partial u_x}{\partial y} \\[4pt]
    \tfrac{\partial u_y}{\partial x}   & 1+\tfrac{\partial u_y}{\partial y}
  \end{pmatrix}, \\
  \det(\mathbf{J}_\varphi) > 0
  &\;\Longleftrightarrow\; 
  \text{locally orientation-preserving (no fold)}.
\end{aligned}
\]
We compute $\det(\mathbf{J}_\varphi)$ via central finite differences at every
pixel of every generated frame for all test-set slices ($30$ frames each).

\begin{figure}[htbp]
  \centering
  \includegraphics[width=\linewidth]{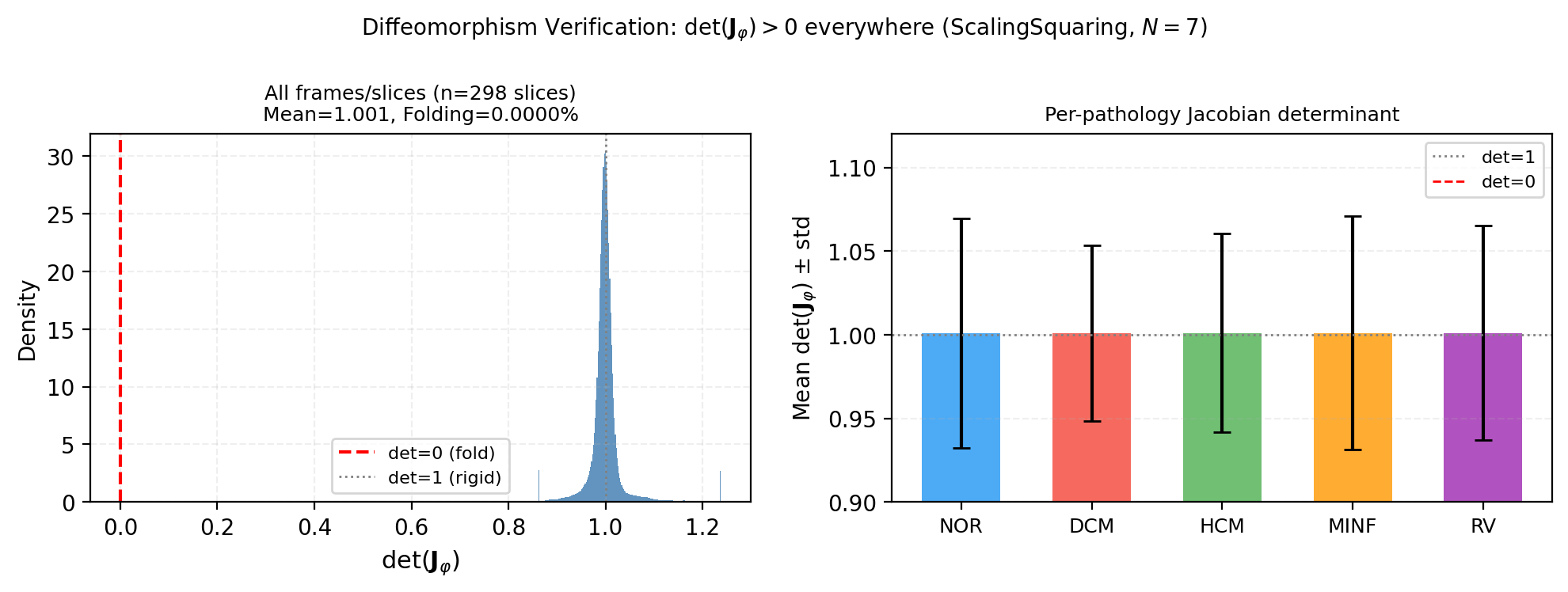}
  \caption{$\det(\mathbf{J}_\varphi)$ distribution (S\&S, $K=7$).
    \emph{Left}: global histogram; red dashed: fold boundary ($\det=0$).
    \emph{Right}: per-pathology mean$\pm$std.
    Zero folding across all pixels confirms diffeomorphic synthesis.}
  \label{fig:jacobian_det}
\end{figure}

\paragraph{Results.}
Figure~\ref{fig:jacobian_det} shows the empirical distribution of
$\det(\mathbf{J}_\varphi)$.
The fraction of pixels with $\det(\mathbf{J}_\varphi)\leq 0$ (folding voxels)
is $\mathbf{0.00\%}$, confirming that S\&S prevents topology violations
across all pathology groups and cardiac phases.
The mean determinant is $1.001\pm0.063$ (mean$\pm$std),
reflecting physiologically expected area changes: myocardial compression during
systole yields $\det<1$ and expansion during diastole yields $\det>1$,
but neither regime produces folds.
Per-pathology statistics (Figure~\ref{fig:jacobian_det}, right) are consistent,
demonstrating that the diffeomorphic property holds across the diverse cardiac
morphologies in ACDC.

\section{Excluded Baselines}
\label{app:baselines}

We detail why four otherwise-natural baselines were excluded from the main
comparison.

\paragraph{DragNet.}
DragNet~\cite{dragnet} synthesizes frames autoregressively, so prediction error
accumulates over the sequence. This drift caused the frozen SegUNet to fail on
19 of 30 test cases (63\%), making the physiological volume-curve evaluation
infeasible.

\paragraph{Registration methods.}
Registration-based methods such as VoxelMorph~\cite{voxelmorph} address a
different task: aligning two \emph{observed} images. They cannot synthesize a
sequence from a single ED frame without a target frame available at inference.

\paragraph{ECG-conditioned methods.}
ECG-conditioned approaches~\cite{echopulse, ecgflowcmr} require concurrent ECG
acquisition, which violates our ECG-free setting and makes direct comparison
unfair.

\paragraph{EchoLVFM.}
EchoLVFM~\cite{echolv}, the concurrent flow-matching method most related to
ours, has no publicly available implementation.

\section{Implementation Details}
\label{app:implementation}

\paragraph{VAE architecture.}
We adopt the 3D VAE architecture from ECGFlowCMR~\cite{ecgflowcmr} with
anisotropic compression: spatial downsampling $8\times$ with temporal dimension
preserved ($f_t = 0$).
Input shape: $(B, 1, T, 128, 128)$; latent shape: $(B, 4, T, 16, 16)$.
We replace the full 3D attention in ECGFlowCMR with per-frame 2D spatial
attention, reducing the sequence length from ${\sim}7{,}680$ to $256$ per
attention block, significantly more memory-efficient and better suited to
the ACDC dataset size.
The VAE is pretrained for 100 epochs (lr = $10^{-4}$, KL weight = $10^{-6}$)
and frozen during flow matching training.

\paragraph{VelocityUNet.}
A 3D UNet with channel multipliers $[1, 2]$, base channels $128$,
temporal self-attention across frames at both resolution levels.
Flow time $\tau$ is injected via sinusoidal embeddings processed by a small MLP,
added to the global condition and applied as scale/shift in adaptive group
normalization (AdaGN) inside each residual block.
The slice-position embedding $\mathbf{e}_{\mathrm{slice}}$ is similarly fused
into AdaGN as a global condition, while the per-frame phase embedding
$\mathbf{e}_{\mathrm{phase}}$ is injected at every resolution level via
cross-attention with $n_{\mathrm{tokens}}=4$ key-value tokens and $n_{\mathrm{heads}}=8$.

\paragraph{Deformation decoder.}
A stack of transposed 2D convolutions (temporal dimension folded into the batch) upsampling from $(4, 16, 16)$ to
$(2, 128, 128)$, producing a 2-channel velocity field (horizontal and
vertical displacement).
Scaling-and-squaring with $K = 7$ steps produces the final diffeomorphic
displacement.
A learnable output scale parameter (initialized to 10.0) controls the
displacement magnitude.

\paragraph{SegUNet.}
A standard 2D UNet~\cite{ronneberger2015unet, isensee2021nnunet} trained on ACDC images with segmentation annotations for the LV, right ventricle (RV), and myocardium.
Used frozen for: (i) nonlinear phase estimation, (ii) VolumeCurveLoss
computation, and (iii) volume-curve evaluation.
The network is never exposed to generated images during training.

\paragraph{Training details.}
All flow matching models are trained for 200 epochs with AdamW
($\beta_1 = 0.9$, $\beta_2 = 0.999$, weight decay $10^{-4}$), learning rate $10^{-4}$ with cosine
annealing, batch size 4, mixed-precision (AMP) training.
LNCC is computed in float32 to avoid numerical cancellation under AMP.
Noise scale $\alpha = 0.1$ for $\mathbf{z}_0$ construction.
For reproducibility, all stochastic components (the train/validation/test
split, network weight initialization, and the anchor noise
$\boldsymbol{\varepsilon}$) are governed by a single fixed random seed ($42$).
Every number reported in this paper (main comparison, ablations, and
appendix analyses) is obtained from a \emph{single} training run under this
fixed seed; we do not average over multiple seeds.

\paragraph{Hyperparameter selection.}
The loss weights were selected on the validation split. We swept the
volume-curve weight $\lambda_{\mathrm{vol}} \in \{0.1, 1.0, 2.0\}$ and the
bending-energy weight $\lambda_{\mathrm{bend}} \in \{0.01, 0.001\}$, and
retained $\lambda_{\mathrm{vol}} = 2.0$ and $\lambda_{\mathrm{bend}} = 0.001$
as the setting that best balanced physiological fidelity (validation
volume-curve $R^2$) against image quality (validation SSIM).
The remaining weights were fixed a priori:
$\lambda_{\mathrm{flow}} = \lambda_{\mathrm{lncc}} = 1.0$, the anchor noise
scale $\alpha = 0.1$, and the scaling-and-squaring steps $K = 7$.
The learning rate ($10^{-4}$), batch size ($4$), and epoch budget ($200$)
followed standard practice for latent flow-matching training and were not
separately tuned.

\paragraph{Computing infrastructure.}
All models were trained and evaluated on a single NVIDIA RTX A6000 GPU
($48$\,GB) under Debian GNU/Linux 11 (bullseye), NVIDIA driver 560.35.03.
The software stack comprised Python 3.9.2, PyTorch 2.7.1 (built against
CUDA 11.8 and cuDNN 9.1), NumPy 2.0.2, and SciPy 1.13.1.

\section{Data Preprocessing}
\label{app:preprocessing}

We describe the full preprocessing pipeline applied to the ACDC cine CMR
volumes. Two offline stages produce (i)~a pathology-stratified patient split
and (ii)~a per-slice nonlinear phase cache; a lightweight per-sample transform
is then applied on the fly inside the \texttt{Dataset}. Algorithm~\ref{alg:preprocess}
gives the exact sequence of operations.

\begin{algorithm}[h]
\caption{ACDC preprocessing (per (patient, slice) sample)}
\label{alg:preprocess}
\begin{algorithmic}[1]
\STATE \textbf{Offline (once):} collect all $150$ patients from
  \texttt{training/} and \texttt{testing/}; read pathology \texttt{Group}
  from each \texttt{Info.cfg}; stratified split by group into
  train/val/test $=105/15/30$ with seed $42$ $\to$ \texttt{split.json}.
\STATE \textbf{Offline (once):} for each (patient, slice), run frozen SegUNet
  to obtain LV area curve; convert to a normalized nonlinear phase curve
  $\phi_t$; store $\to$ \texttt{phase\_nonlinear.json}.
\STATE \textbf{Per sample} $(\text{patient}, z)$:
\STATE \quad Load 4D volume $(H,W,Z,T_{\mathrm{orig}})$; take slice $z$ $\to$
  $(T_{\mathrm{orig}}, H, W)$.
\STATE \quad Read ED/ES frame indices from \texttt{Info.cfg} (ED $=$ frame $0$).
\STATE \quad Intensity: per-sample min--max normalize to $[-1,1]$.
\STATE \quad Spatial: bilinear resize each frame to $128\times128$.
\STATE \quad Temporal: trilinear resample to $T=30$ frames; resample $\phi_t$
  and rescale the ES index accordingly.
\STATE \quad Encode phase as $[\sin\phi_t,\cos\phi_t]$; set anchor
  $\mathbf{x}_{\mathrm{ED}}=$ frame $0$; slice position $=z/(Z-1)$.
\STATE \textbf{return} $(\text{cine}, \mathbf{x}_{\mathrm{ED}}, \phi_{\mathrm{seq}},
  \text{slice\_pos}, \text{es\_idx})$.
\end{algorithmic}
\end{algorithm}

No intensity augmentation, cropping, or registration is applied; the only
spatial operation is an isotropic resize. Intensity normalization is performed
per (patient, slice) sequence so that all frames of a given slice share one
$[\,v_{\min}, v_{\max}\,]$ range, preserving relative temporal contrast.
The stratified split and the phase cache are computed once and reused across
all models (main and baselines) to ensure an identical evaluation protocol.

\section{Evaluation Metrics}
\label{app:metrics}

We formally define the eight metrics used in the main paper and state the
motivation for each. Let $\hat{\mathbf{x}}_t$ and $\mathbf{x}_t$ denote the
generated and ground-truth frame at time $t$ ($t=1,\dots,T$), with pixel
intensities rescaled to $[0,1]$.

\paragraph{Image quality.}
\begin{itemize}\setlength{\itemsep}{.2em}
  \item \textbf{PSNR} (dB), averaged over frames:
    $\mathrm{PSNR}=10\log_{10}\!\big(1/\mathrm{MSE}(\hat{\mathbf{x}}_t,\mathbf{x}_t)\big)$.
    Motivation: standard pixel-level fidelity measure; penalizes reconstruction
    error directly.
  \item \textbf{SSIM}~\cite{wang2004ssim}:
    $\mathrm{SSIM}=\frac{(2\mu_{\hat{x}}\mu_x+c_1)(2\sigma_{\hat{x}x}+c_2)}
    {(\mu_{\hat{x}}^2+\mu_x^2+c_1)(\sigma_{\hat{x}}^2+\sigma_x^2+c_2)}$.
    Motivation: captures structural/luminance/contrast agreement, correlating
    better with perceived quality than PSNR for anatomical structure.
  \item \textbf{LPIPS}~\cite{zhang2018lpips}: distance between deep features of
    $\hat{\mathbf{x}}_t$ and $\mathbf{x}_t$.
    Motivation: perceptual similarity that is sensitive to texture realism,
    which pixel metrics miss.
  \item \textbf{FID}~\cite{heusel2017fid}:
    $\lVert\mu_r-\mu_g\rVert^2+\mathrm{Tr}\!\big(\Sigma_r+\Sigma_g
    -2(\Sigma_r\Sigma_g)^{1/2}\big)$ over Inception features of real ($r$) and
    generated ($g$) frames.
    Motivation: measures distributional realism of the generated frames as a
    set, complementing per-frame metrics.
\end{itemize}

\paragraph{Physiological fidelity.}
Let $V_t$ and $\hat{V}_t$ be the ground-truth and predicted LV cavity volume at
frame $t$. The predicted volume $\hat{V}_t$ is obtained by warping the ED
segmentation mask through the predicted displacement field, so the segmentation
network is never applied to generated frames; the ground-truth volume $V_t$ is
obtained by segmenting the corresponding real frame $\mathbf{x}_t$.
\begin{itemize}\setlength{\itemsep}{.2em}
  \item \textbf{EF~MAE} (\%): absolute error of ejection fraction
    $\mathrm{EF}=(V_{\max}-V_{\min})/V_{\max}$.
    Motivation: EF is the primary clinical index of cardiac function; a direct
    measure of physiological correctness.
  \item \textbf{Vol~Corr}: Pearson correlation between $\{\hat{V}_t\}$ and
    $\{V_t\}$.
    Motivation: tests whether the \emph{shape} of the volume--time trajectory
    is reproduced, independent of scale.
  \item \textbf{Vol~$R^2$}: coefficient of determination
    $1-\sum_t(\hat{V}_t-V_t)^2/\sum_t(V_t-\bar{V})^2$.
    Motivation: penalizes both bias and variance of the volume curve; negative
    values expose predictions worse than the mean.
  \item \textbf{Vol~MAE}: mean absolute per-frame volume error, normalized by
    ED volume.
    Motivation: a scale-aware complement to the correlation-based measures.
\end{itemize}
Physiological metrics are computed on warped ED masks rather than on
segmentations of the generated frames, avoiding segmentation domain shift in
the evaluation pipeline.

\section{Statistical Analysis}
\label{app:statistics}

\paragraph{Metric variability.}
All reported numbers are obtained from a single training run under fixed seed
$42$; to characterize the distribution of performance beyond point estimates,
Table~\ref{tab:variability} reports the per-case mean and standard deviation of
PhaseFlow across the test set. Image-quality metrics are computed per slice
($n{=}298$) and physiological metrics per patient ($n{=}30$).
The spread is small for image quality (e.g.\ SSIM $0.956\pm0.019$) but larger
for physiological metrics (e.g.\ Vol~$R^2$ $0.363\pm0.489$), reflecting genuine
between-patient variation in cardiac motion and pathology severity rather than
estimator noise.

\begin{table}[h]
  \centering
  \small
  \begin{tabular}{lc}
    \toprule
    Metric & Mean $\pm$ Std \\
    \midrule
    \multicolumn{2}{l}{\emph{Image quality (per slice, $n{=}298$)}} \\
    \quad PSNR (dB)        & $31.72 \pm 3.33$ \\
    \quad SSIM             & $0.956 \pm 0.019$ \\
    \quad LPIPS            & $0.023 \pm 0.012$ \\
    \midrule
    \multicolumn{2}{l}{\emph{Physiological fidelity (per patient, $n{=}30$)}} \\
    \quad EF MAE (\%)      & $17.79 \pm 11.15$ \\
    \quad Vol Corr         & $0.867 \pm 0.132$ \\
    \quad Vol $R^2$        & $0.363 \pm 0.489$ \\
    \quad Vol MAE          & $0.101 \pm 0.060$ \\
    \bottomrule
  \end{tabular}
  \caption{Per-case mean $\pm$ standard deviation of PhaseFlow on the ACDC test
    set. Image metrics are per-slice ($n{=}298$); physiological metrics are
    per-patient ($n{=}30$).}
  \label{tab:variability}
\end{table}

\paragraph{Significance testing.}
To verify that PhaseFlow's physiological improvements are statistically
significant rather than artifacts of point estimates, we apply the paired
Wilcoxon signed-rank test on the per-patient physiological metrics ($n{=}30$)
against four baselines spanning the trivial null (ED~Repeat), a recurrent
predictor (ConvLSTM), and two strong learned generators (CVAE, Direct
Registration).
Table~\ref{tab:wilcoxon} reports the resulting $p$-values.
Each cell is a $p$-value, \emph{not} a performance score: it measures how
unlikely the observed PhaseFlow-vs-baseline gap would be if the two methods were
truly equal, so a \emph{smaller} value is \emph{stronger} evidence in favour of
PhaseFlow. Bold entries are significant at $\alpha{=}0.05$.

The central result is Vol~$R^2$: PhaseFlow is significantly better than
\emph{every} baseline ($p$ ranging from $1.6{\times}10^{-2}$ to
$1.9{\times}10^{-9}$), confirming the main-text finding that it is the only
method to reproduce ventricular contraction \emph{magnitude} (positive $R^2$).
PhaseFlow is likewise significantly better on Vol~MAE and EF~MAE against all
baselines except Direct Registration.
On Vol~Corr, which measures only the \emph{shape} (timing) of the volume curve
and not its magnitude, PhaseFlow is statistically indistinguishable from the two
strongest baselines (CVAE and Direct Registration; $p{=}0.36$ and $0.72$):
these baselines track \emph{when} the ventricle contracts but, as their negative
$R^2$ shows, not by \emph{how much}, which is precisely the failure mode the
VolumeCurveLoss is designed to correct.
The Vol~Corr comparison against ED~Repeat is undefined (the static ED~Repeat
sequence has a constant, zero-variance volume curve, so Pearson correlation
cannot be computed) and is omitted.

\begin{table}[h]
  \centering
  \small
  \setlength{\tabcolsep}{4pt}
  \begin{tabular}{lcccc}
    \toprule
    & \multicolumn{4}{c}{Wilcoxon $p$ (PhaseFlow vs.)} \\
    \cmidrule(lr){2-5}
    Metric & ED~Rep. & ConvLSTM & CVAE & Dir.~Reg. \\
    \midrule
    EF MAE    & $\mathbf{1.9{\times}10^{-9}}$ & $\mathbf{1.9{\times}10^{-9}}$ & $\mathbf{2.0{\times}10^{-2}}$ & $0.36$ \\
    Vol Corr  & ---                          & $\mathbf{5.6{\times}10^{-9}}$ & $0.36$                       & $0.72$ \\
    Vol $R^2$ & $\mathbf{1.9{\times}10^{-9}}$ & $\mathbf{6.2{\times}10^{-8}}$ & $\mathbf{2.3{\times}10^{-4}}$ & $\mathbf{1.6{\times}10^{-2}}$ \\
    Vol MAE   & $\mathbf{5.6{\times}10^{-9}}$ & $\mathbf{2.6{\times}10^{-8}}$ & $\mathbf{2.3{\times}10^{-4}}$ & $0.064$ \\
    \bottomrule
  \end{tabular}
  \caption{Paired Wilcoxon signed-rank test ($n{=}30$ patients) against four
    baselines. Every cell is a $p$-value; \emph{smaller means stronger evidence
    that PhaseFlow is better}. \textbf{Bold} $=$ significant at $\alpha{=}0.05$.
    PhaseFlow is significantly better on Vol~$R^2$ against \emph{all} baselines;
    non-bold cells (Vol~Corr vs.\ CVAE/Dir.~Reg., EF/Vol~MAE vs.\ Dir.~Reg.)
    indicate no significant difference. Vol~Corr vs.\ ED~Repeat is undefined
    (constant volume curve).}
  \label{tab:wilcoxon}
\end{table}

\section{Latent Scale and Anchor Perturbation}
\label{app:latent_scale}

The anchor noise scale $\alpha$ in the anchor latent construction $\mathbf{z}_0 = E_{\mathrm{vae}}(\mathrm{repeat}(\mathbf{x}_{\mathrm{ED}},\, T)) + \alpha\boldsymbol{\varepsilon}$ (with $\boldsymbol{\varepsilon} \sim \mathcal{N}(\mathbf{0},\,\mathbf{I})$) is only meaningful
relative to the scale of the latent space itself. We therefore measure the
empirical statistics of the frozen VAE latents over the ACDC splits
(Table~\ref{tab:latent_scale}). Latents are taken as the posterior mode
$\boldsymbol{\mu}$, matching the training configuration; the latent tensor has
dimension $D = C \times T \times h \times w = 4 \times 30 \times 16 \times 16 =
30{,}720$.

\begin{table}[h]
  \centering
  \small
  \begin{tabular}{lcc}
    \toprule
    Quantity & Train ($n{=}1048$) & Val ($n{=}143$) \\
    \midrule
    $\sigma$ of $\mathbf{z}_1$                        & 0.506 & 0.538 \\
    $\sigma$ of anchor $E_{\mathrm{vae}}(\mathrm{repeat}(\cdot))$ & 0.503 & 0.535 \\
    $\sigma$ of $\mathbf{z}_1 - \mathbf{z}_0$         & 0.171 & 0.174 \\
    \midrule
    $\alpha / \sigma(\mathbf{z}_1)$                   & 19.8\% & 18.6\% \\
    $\alpha / \sigma(\mathbf{z}_1 - \mathbf{z}_0)$    & 58.3\% & 57.4\% \\
    \bottomrule
  \end{tabular}
  \caption{Empirical latent statistics ($\sigma$ = per-element standard
    deviation). The two splits agree closely, indicating the latent scale is a
    stable property of the frozen encoder rather than a split artifact.}
  \label{tab:latent_scale}
\end{table}

\paragraph{Geometry of the perturbation.}
The per-element ratios in Table~\ref{tab:latent_scale} (up to $58\%$) make the
anchor noise look large, but they \emph{overstate} its effect on the motion
signal. In the high-dimensional latent space the noise is almost entirely
orthogonal to the motion direction: it corrupts the true transport direction by
only $0.28\%$. The anchor perturbation is therefore effectively harmless to the
motion content, and $\alpha{=}0.1$ is a safe choice.

Concretely, on the validation split the transport distance is
$\lVert \mathbf{z}_1 - \mathbf{z}_0 \rVert = 29.33 \pm 8.51$ while the noise
norm is $\lVert \alpha\boldsymbol{\varepsilon} \rVert = 17.54 \pm 0.07$
(matching the theoretical $\alpha\sqrt{D} = 17.53$), a seemingly large $66\%$
ratio. This ratio is misleading because the noise energy is spread across all
$D = 30{,}720$ dimensions: its projection onto the single transport direction
$\mathbf{z}_1 - \mathbf{z}_0$ is only $0.074 \pm 0.089$, i.e.\ $0.284\%$ of the
transport distance, matching the theoretical prediction that for isotropic
$\boldsymbol{\varepsilon}$ this projection follows $\mathcal{N}(0, \alpha^2)$
independently of $D$ ($\mathbb{E}|{\cdot}| = \alpha\sqrt{2/\pi} = 0.080$).
The perturbed and clean transport directions differ by only
$32.5^\circ \pm 8.3^\circ$: the noise samples a wide cone of starting points
around the anchor, but all such paths terminate at the same $\mathbf{z}_1$, so
the component that actually corrupts the motion signal is negligible.

\section{Extended Qualitative Results}
\label{app:qualitative}

\begin{figure}[htbp]
  \centering
  \includegraphics[width=\linewidth]{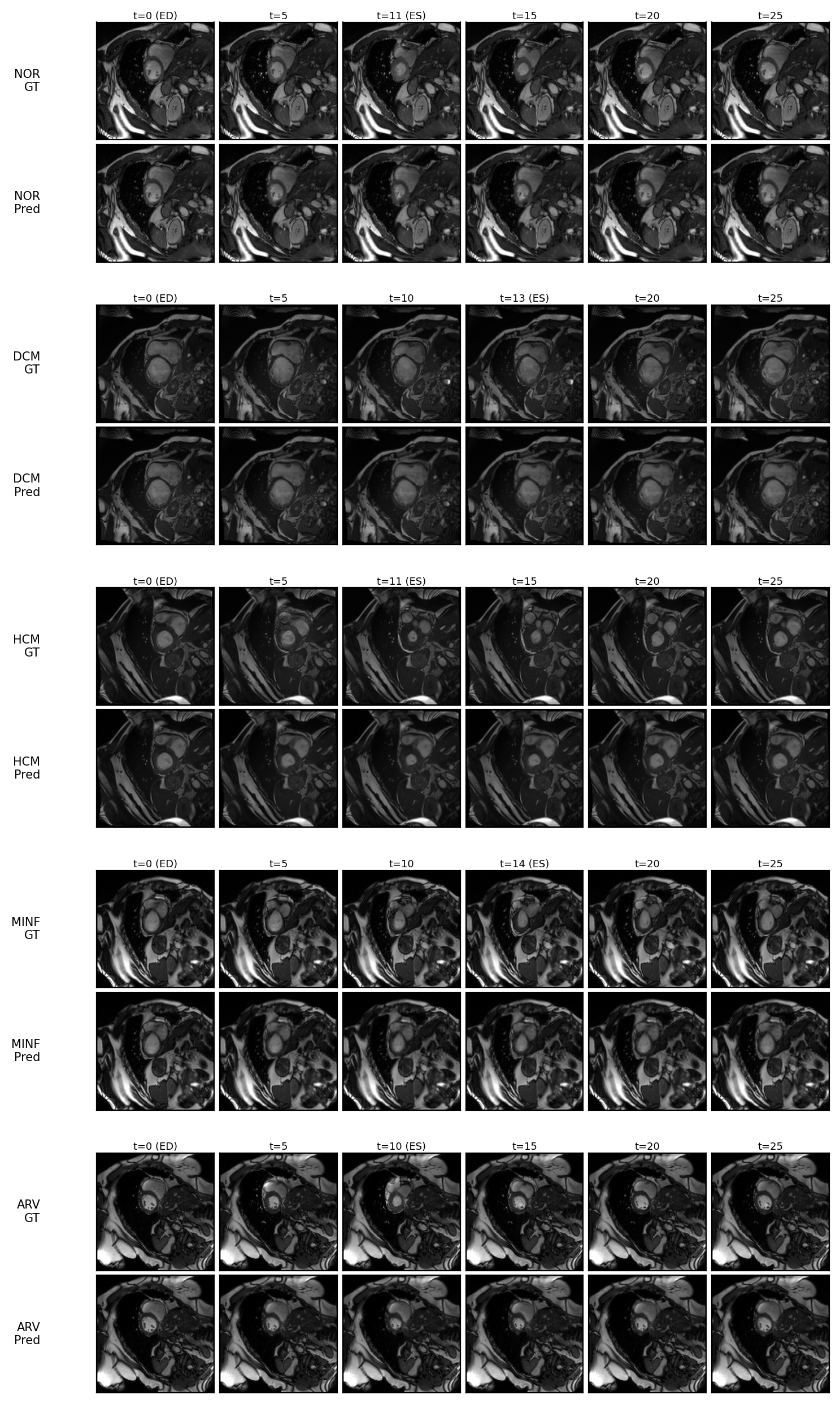}
  \caption{Per-pathology qualitative results (one representative patient per group).
    \emph{Top row}: ground truth; \emph{bottom row}: PhaseFlow synthesis.
    ED frame ($t=0$, leftmost) is the sole model input.}
  \label{fig:qualitative_grid}
\end{figure}

Figure~\ref{fig:qualitative_grid} presents representative generated cardiac
cine sequences for one patient from each ACDC pathology group.
For each group, a representative patient is manually selected to clearly
illustrate the characteristic motion signature of that pathology.

Each group occupies two rows: the top row shows the ground-truth sequence and
the bottom row shows the PhaseFlow-generated sequence, at six evenly-spaced
frames spanning the full cardiac cycle.
The leftmost frame (ED, $t=0$) is the only input to the model at inference
time; all subsequent frames are synthesised from the nonlinear phase template
and the predicted diffeomorphic displacement field.

PhaseFlow faithfully captures the characteristic motion signatures of each
pathology: normal wall-thickening in NOR, large-cavity contraction in DCM,
asymmetric septal hypertrophy in HCM, regional wall-motion abnormality in MINF,
and preserved right ventricular dynamics in ARV.

\end{document}